\documentclass[10pt]{IEEEtran}
\IEEEoverridecommandlockouts
\usepackage{cite}
\usepackage{amsmath,amssymb,amsfonts}
\usepackage{graphicx}
\usepackage{textcomp}
\usepackage{amsmath}
\usepackage{booktabs} 
\usepackage{multirow}
\usepackage{booktabs}
\usepackage{tabularx}
\usepackage{array}
\usepackage{hyperref}       
\usepackage{url}            
\usepackage{booktabs}       
\usepackage{amsfonts}       
\usepackage{nicefrac}       
\usepackage{microtype}      
\usepackage{xcolor}         
\usepackage{setspace}
\usepackage{algpseudocode}
\makeatletter
\let\STATE\State
\let\REQUIRE\Require
\let\ENSURE\Ensure
\let\FOR\For
\let\ENDFOR\EndFor

\newcommand{\TO}{\textbf{to}}

\newcommand{\RETURN}[1]{\State \Return #1}
\makeatother

\usepackage{amsmath}
\usepackage{tikz}

\usepackage{orcidlink}
\usepackage{comment}
\usepackage{amssymb}
\usepackage{braket}
\usepackage{xcolor}
\usepackage[font=footnotesize,skip=0pt]{caption}
\usepackage{booktabs}
\usepackage{multirow}
\usepackage{adjustbox}
\usepackage{array}
\usepackage{graphicx} 
\usepackage[table]{xcolor}
\usepackage{algorithm}
\usepackage{algpseudocode} 
\makeatletter
\let\COMMENT\Comment
\makeatother

\newcommand{\best}[1]{\textbf{\cellcolor{green!15}#1}}
\def\BibTeX{{\rm B\kern-.05em{\sc i\kern-.025em b}\kern-.08em
    T\kern-.1667em\lower.7ex\hbox{E}\kern-.125emX}}

    \usepackage{fancyhdr}
\fancypagestyle{firstpage}{
   \fancyhf{} 
   \fancyhead[C]{To appear at 2026 IEEE International Joint Conference on Neural Networks (IJCNN) Proceedings, Maastricht, Netherlands} 
}
\begin{document}

\title{
How Embeddings Shape Graph Neural Networks: Classical vs Quantum-Oriented Node Representations}

\author{\IEEEauthorblockN{Nouhaila Innan\textsuperscript{1,2}, Antonello Rosato\textsuperscript{3}, Alberto Marchisio\textsuperscript{1,2}, and Muhammad Shafique\textsuperscript{1,2}\\}
\IEEEauthorblockA{
\textsuperscript{1}eBRAIN Lab, Division of Engineering, New York University Abu Dhabi (NYUAD), Abu Dhabi, UAE\\
\textsuperscript{2}Center for Quantum and Topological Systems (CQTS), NYUAD Research Institute, NYUAD, Abu Dhabi, UAE\\
\textsuperscript{3}Department of Information Engineering, Electronics and Telecommunications, Sapienza University of Rome, Rome, Italy\\
\{nouhaila.innan, alberto.marchisio, muhammad.shafique\}@nyu.edu, antonello.rosato@uniroma1.it\\
}}

\maketitle
\thispagestyle{empty} 
\thispagestyle{firstpage}
\begin{abstract}
Node embeddings act as the information interface for graph neural networks, yet their empirical impact is often reported under mismatched backbones, splits, and training budgets. This paper provides a controlled benchmark of embedding choices for graph classification, comparing classical baselines with quantum-oriented node representations under a unified pipeline. We evaluate two classical baselines alongside quantum-oriented alternatives, including a circuit-defined variational embedding and quantum-inspired embeddings computed via graph operators and linear-algebraic constructions. All variants are trained and tested with the same backbone, stratified splits, identical optimization and early stopping, and consistent metrics. Experiments on five different TU datasets and on QM9 converted to classification via target binning show clear dataset dependence: quantum-oriented embeddings yield the most consistent gains on structure-driven benchmarks, while social graphs with limited node attributes remain well served by classical baselines. The study highlights practical trade-offs between inductive bias, trainability, and stability under a fixed training budget, and offers a reproducible reference point for selecting quantum-oriented embeddings in graph learning.
\end{abstract}

\begin{IEEEkeywords}
Graph Neural Networks, Embeddings, Positional Encoding. 
\end{IEEEkeywords}

\section{Introduction}

Graph-level prediction depends strongly on how node information is represented before and during message passing. Although node embeddings can substantially affect the information available to the graph classifier, their contribution is often difficult to assess in isolation because prior comparisons are frequently confounded by differences in backbone architectures, data splits, optimization settings, and training budgets.

This work studies, under a unified protocol, how different embedding modules influence downstream graph-level prediction when the GNN backbone is kept fixed. Specifically, we integrate a family of quantum-inspired embedding constructions into the same GNN pipeline and evaluate them on a shared set of graph classification benchmarks, including a molecular benchmark. Performance is measured using test Accuracy, Macro-F1, and Macro Precision/Recall. Rather than claiming that one embedding family is universally superior, the aim is to isolate the effect of the embedding stage and identify when different inductive biases are beneficial.

Across all experiments, each node is described using lightweight structural and positional information. Let $x$ denote the base node feature vector, constructed from one-hot degree encodings, and let $pe$ denote the Laplacian positional encoding (LPE) vector computed from the eigenvectors of the normalized graph Laplacian \cite{dwivedi2022graph}. The notation $(x \Vert pe)$ denotes the concatenation of these two vectors. Some methods operate directly on this representation, while others generate additional features through graph-dependent dynamics and fuse them with the same base input.

The embedding variants considered in this paper cover a range of design principles. As classical controls, we include a \textbf{Fixed} embedding obtained through a fixed random projection of $(x \Vert pe)$ and a trainable multi-layer perceptron (\textbf{MLP}) applied to $(x \Vert pe)$. As a trainable quantum-circuit baseline, we include a variational quantum circuit (\textbf{VQC}) with angle encoding, denoted as \textbf{Angle-VQC}, where node features are encoded as circuit angles and measurement outcomes define node embeddings \cite{cerezo2021variational,innan2024financial}. We then consider three quantum-inspired constructions defined through graph dynamics:

\begin{itemize}
    \item \textbf{QuOp}, which derives embeddings from local operator evolution on ego-neighborhood subgraphs,
    \item \textbf{QWalkVec}, which constructs node vectors from coined quantum-walk-style evolution over a fixed number of steps,
    \item \textbf{QPE}, which produces node descriptors from transition probabilities derived from a matrix-exponential formulation with anchor-node conditioning.
\end{itemize}
For both QuOp and QWalkVec, we evaluate non-trainable and trainable versions. In the results, the \emph{trainable} variants are marked with a superscript $^{*}$, indicating that a learnable fusion or projection stage is enabled.

These methods were selected to represent different ways of constructing node embeddings, including projection-based baselines, circuit-defined quantum embeddings, operator-based encodings, walk-based encodings, and spectral or positional constructions. Comparing them within one shared pipeline allows us to examine whether these different inductive biases lead to different graph-classification behavior under matched training conditions.

The central question of this study is: \emph{given the same GNN backbone and training protocol, when does injecting operator- or walk-based structure into the embedding stage improve graph-level prediction relative to standard projection-based or MLP-based alternatives?} To answer this, we vary only the embedding module while keeping the downstream architecture and training setup unchanged. We also compare trainable and non-trainable variants to distinguish gains arising from additional learnable capacity from those due to the embedding construction itself. Our working hypothesis is that embeddings encoding higher-order structure, walk dynamics, or positional information may be more effective when graph labels depend on multi-hop structural patterns, whereas simpler baselines may remain competitive when local descriptors are already sufficient.

The main contribution of this paper is a unified experimental framework for evaluating the role of the node embedding stage under a fixed GNN backbone and shared protocol. By reporting multiple metrics and, where applicable, both trainable and non-trainable configurations, the study clarifies when performance differences are attributable to the embedding design itself and when they are associated with additional trainable components. Because all methods are evaluated under the same finite optimization budget, the benchmark should be interpreted as a comparison under equal training resources rather than as an estimate of the maximum achievable performance of each method after exhaustive tuning.

The remainder of the paper presents the embedding constructions and their integration into the shared model, followed by an empirical analysis on TU \cite{morris2020tudataset} and QM9 \cite{ramakrishnan2014qm9} datasets, and a discussion of the methodological insights suggested by the observed performance patterns.

\section{Background and Related Work}
\label{sec:related}

\subsection{Node representations for graph classification}
Graph neural networks (GNNs) rely on node-level inputs that determine what information can be propagated and pooled at the graph level \cite{kipf2017semi}. 
In practical benchmarks, node attributes may be rich (molecular graphs) or entirely absent (social-network TU datasets such as IMDB variants).
Graph benchmarking is a specific challenging problem \cite{dwivedi2020benchmarking}, and when attributes are missing, a common strategy is to construct simple structural features (e.g., node degree) to provide a consistent input interface, while keeping the injected signal minimal \cite{velivckovic2018graph}. 
In this work, we focus on how different node-embedding mechanisms, including quantum-oriented representations, affect downstream graph classification when the backbone and training protocol are held fixed.

\subsection{Quantum-oriented node embeddings}
Quantum-oriented embeddings fall into two broad families. 
The first family is \emph{circuit-defined embeddings}, where a parameterized quantum circuit maps node features to a fixed-dimensional representation (e.g., Angle-VQC). 
The second family is \emph{quantum-inspired embeddings}, which borrow primitives from quantum dynamics or quantum information (operators, walks, phase encodings) but are computed via graph operators and linear algebra \cite{aharonov2001quantumwalks}. 
Our benchmark includes three representative quantum-inspired methods from recent literature: QuOp \cite{vlasic2025quop}, QWalkVec \cite{sato2024qwalkvec}, and QPE \cite{thabet2024quantum}. 
While these approaches are motivated by quantum formalisms, in our implementation, they are evaluated under the same classical training pipeline and compared against classical baselines.

\subsubsection{QuOp: quantum operator node representations}
QuOp \cite{vlasic2025quop} constructs a node representation from local operator dynamics on an ego neighborhood. 
For each node, a neighborhood-induced operator is mapped to a unitary transformation, and summary statistics of the resulting state are used as the embedding. 
This yields a representation that explicitly encodes local graph structure while controlling dimensionality through a qubit budget. 
In our benchmark, we evaluate both a non-trainable variant (QuOp) and a trainable variant (QuOp*), where the trainable version introduces a learned projection that maps the operator-derived descriptors into the shared embedding dimension used by the downstream GIN (see Algorithm \ref{alg:quop}).

\begin{algorithm}[htpb]
\caption{QuOp node embedding}
\label{alg:quop}
\small
\begin{algorithmic}[1]
\REQUIRE Graph $G=(V,E)$, node $v$, hop radius $h$, qubit budget $q$, output dim $d$
\ENSURE Node embedding $\mathbf{z}_v \in \mathbb{R}^d$
\STATE Extract the $h$-hop ego-subgraph $G_v$ centered at node $v$
\STATE Construct a graph-dependent matrix $H_v$ from the local structure of $G_v$
\STATE Resize $H_v$ to dimension $2^q \times 2^q$ to match the $q$-qubit state space
\STATE Compute the unitary matrix $U_v \leftarrow \exp(-i\tilde{H}_v)$
\STATE Initialize a basis state $|\psi_0\rangle$ associated with the target node
\STATE Evolve the state: $|\psi_v\rangle \leftarrow U_v |\psi_0\rangle$
\STATE Extract a fixed summary vector $\mathbf{s}_v$ from the evolved state $|\psi_v\rangle$
\STATE Map $\mathbf{s}_v$ to the shared embedding dimension: $\mathbf{z}_v \leftarrow \textsc{Proj}(\mathbf{s}_v)$ \COMMENT{identity for QuOp; learnable for QuOp*}
\RETURN $\mathbf{z}_v$
\end{algorithmic}
\end{algorithm}

\subsubsection{QWalkVec: node embeddings from quantum-walk dynamics}
QWalkVec \cite{sato2024qwalkvec} constructs node embeddings by simulating a coined quantum walk and recording how node visitation probabilities evolve over time. In this way, each node is represented by a temporal descriptor that reflects how information propagates through multi-hop graph structure under the walk dynamics. We benchmark a fixed variant (QWalkVec) and a trainable variant (QWalkVec*), where the latter learns a projection from these time-series probability descriptors to the shared embedding dimension used downstream. Empirically, this projection can be important for aligning the walk-derived signal with the supervised task (see Algorithm \ref{alg:qwalkvec}).

\begin{algorithm}[htpbt]
\caption{QWalkVec}
\label{alg:qwalkvec}
\small
\begin{algorithmic}[1]
\REQUIRE Graph $G=(V,E)$, steps $T$, walk parameters $(w_p,w_q)$, output dim $d$
\ENSURE Node embeddings $\{\mathbf{z}_v\}_{v\in V}$, $\mathbf{z}_v \in \mathbb{R}^d$
\STATE Construct the directed-edge state space $\mathcal{S} \leftarrow \{(u\!\rightarrow\!v):(u,v)\in E\}$
\STATE Initialize the walk state $|\Psi^{(0)}\rangle$ on $\mathcal{S}$
\FOR{$t=1$ \TO $T$}
  \STATE Apply the coin and shift updates:
  \STATE \hspace{1em} $|\Psi^{(t)}\rangle \leftarrow S \, C(w_p,w_q) \, |\Psi^{(t-1)}\rangle$
  \STATE For each node $v$, compute the node visitation probability $p_v^{(t)}$ by summing the probabilities of all directed-edge states in $|\Psi^{(t)}\rangle$ that end at or are associated with $v$
\ENDFOR
\FOR{each node $v\in V$}
  \STATE Form the time-series descriptor $\mathbf{s}_v \leftarrow [p_v^{(1)},\dots,p_v^{(T)}]$
  \STATE Map $\mathbf{s}_v$ to the shared embedding dimension: $\mathbf{z}_v \leftarrow \textsc{Proj}(\mathbf{s}_v)$ \COMMENT{fixed for QWalkVec; learnable for QWalkVec*}
\ENDFOR
\RETURN $\{\mathbf{z}_v\}_{v\in V}$
\end{algorithmic}
\end{algorithm}

\subsubsection{QPE: quantum positional encodings for graphs}
QPE \cite{thabet2024quantum} injects positional information into node representations through quantum-inspired phase dynamics. The main idea is to encode node position through the spectral structure of a graph-derived operator, such as a Laplacian-based matrix, and to evaluate the time evolution $U(t)=\exp(-iHt)$ at multiple time points. This produces node descriptors that reflect how each node responds to the same operator dynamics relative to a selected set of anchor nodes. In our benchmark, QPE is implemented as a quantum-inspired linear-algebra construction and evaluated under the same fixed GIN backbone (see Algorithm \ref{alg:qpe}).

\begin{algorithm}[htpbt]
\caption{QPE}
\label{alg:qpe}
\small
\begin{algorithmic}[1]
\REQUIRE Graph $G=(V,E)$, operator $H$, time set $\mathcal{T}$, number of anchors $A$, output dim $d$
\ENSURE Node embeddings $\{\mathbf{z}_v\}_{v\in V}$, $\mathbf{z}_v \in \mathbb{R}^d$
\STATE Select a set of anchor nodes $\mathcal{A} \subseteq V$ with $|\mathcal{A}| = A$
\STATE Compute the spectral decomposition $H = V \Lambda V^\top$
\FOR{each $t \in \mathcal{T}$}
  \STATE Compute the evolution operator $U(t) \leftarrow V \exp(-i\Lambda t) V^\top$
  \FOR{each anchor $a \in \mathcal{A}$}
    \STATE Define the anchor indicator vector $\mathbf{e}_a$
    \STATE Evolve the anchor state: $\mathbf{u}_{a,t} \leftarrow U(t)\mathbf{e}_a$
    \FOR{each node $v \in V$}
      \STATE Append $\mathbf{u}_{a,t}[v]$ to the descriptor list of node $v$
    \ENDFOR
  \ENDFOR
\ENDFOR
\FOR{each node $v \in V$}
  \STATE Concatenate all collected values into a descriptor vector $\mathbf{s}_v$
  \STATE Map $\mathbf{s}_v$ to the shared embedding dimension: $\mathbf{z}_v \leftarrow \textsc{Proj}(\mathbf{s}_v)$ \COMMENT{fixed in our QPE configuration}
\ENDFOR
\RETURN $\{\mathbf{z}_v\}_{v\in V}$
\end{algorithmic}
\end{algorithm}
\section{Methodology}
Our goal is to quantify how the \emph{node embedding stage} affects graph-level prediction under an otherwise fixed pipeline.
Let $G=(V,E)$ denote an input graph with $|V|$ nodes and edge set $E$. Let $\phi_\theta$ denote the node-embedding module producing $\mathbf{Z}\in\mathbb{R}^{|V|\times d}$.

\begin{figure*}
    \centering
    \includegraphics[width=\linewidth]{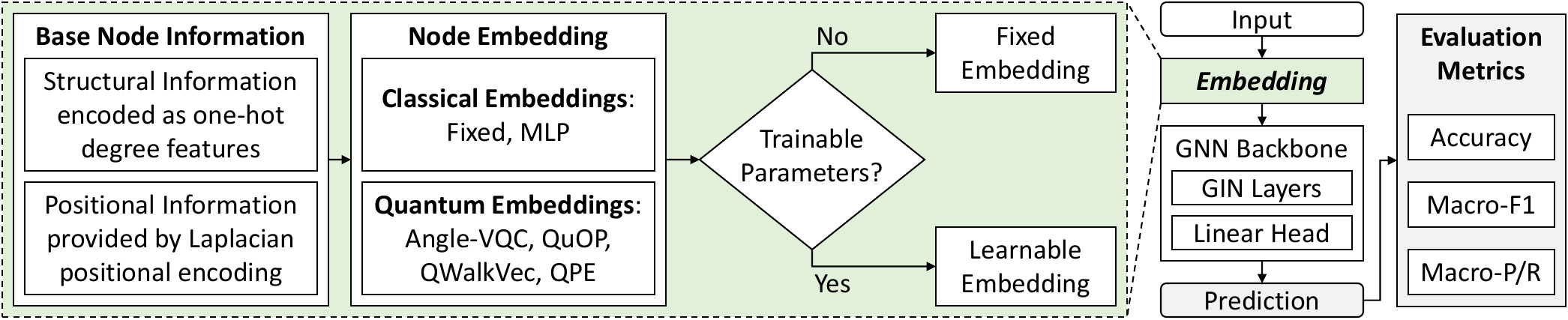}
    \caption{Visual representation of the complete methodology for evaluating embeddings.}
    \label{fig:methodology}
\end{figure*}

The resulting node embeddings are then consumed by the same downstream classifier $f_\psi(\cdot)$ (a fixed GIN backbone) to produce graph logits and predictions.
As shown in Fig.~\ref{fig:methodology}, across all runs, we keep constant: (i) the backbone architecture $f_\psi$, (ii) the data splits, (iii) the optimizer and early-stopping criterion, and (iv) the evaluation metrics.
Therefore, differences in results are attributable to $\phi_\theta$ (and to whether $\phi_\theta$ is trainable), rather than to confounding changes in message passing or the training procedure.
\subsection{Node embedding constructions}
Each embedding method defines a mapping from the graph and per-node descriptors to a fixed-dimensional node representation,
\begin{equation}
\mathbf{Z}=\phi_\theta\!\left(G,\{\mathbf{u}_v\}_{v\in V}\right)\in\mathbb{R}^{|V|\times d},
\label{eq:phi}
\end{equation}
where $d_u$ denotes the dimensionality of the input node descriptor $\mathbf{u}_v$, and $\mathbf{z}_v$ denotes the $v$-th row of $\mathbf{Z}$. For all methods, the embedding stage can be expressed as the composition of a method-specific descriptor and a mapping into the shared space:
\begin{align}
\mathbf{s}_v &= g(G,v;\alpha)\in\mathbb{R}^{d_s}, \label{eq:sv}\\
\mathbf{z}_v &= \rho_\theta\!\left([\mathbf{u}_v \Vert \mathbf{s}_v]\right)\in\mathbb{R}^{d}. \label{eq:zv}
\end{align}
The function $g(\cdot)$ is determined by the chosen embedding construction and fixed hyperparameters $\alpha$ (e.g., walk horizon, anchor count, or qubit budget). The map $\rho_\theta(\cdot)$ projects the resulting descriptor into the shared embedding dimension $d$. In the non-trainable setting, $\rho_\theta$ is fixed; in the trainable setting, $\rho_\theta$ contains learnable parameters. The \textbf{Trainable} indicator used in the results indicates whether the embedding stage introduces learnable parameters beyond the shared GIN backbone (e.g., a learnable projection/fusion map and, for Angle-VQC, circuit parameters). Detailed algorithmic descriptions of the quantum-inspired embeddings (QuOp, QWalkVec, and QPE) are provided in Sec.~\ref{sec:related}.

\paragraph{Classical baselines}
Classical controls operate directly on $\mathbf{u}_v$ by setting $\mathbf{s}_v=\varnothing$ in Eq.~\eqref{eq:zv}. The fixed baseline applies a fixed random projection,
\begin{equation}
\mathbf{z}_v = \mathbf{W}_0\,\mathbf{u}_v,\qquad \mathbf{W}_0\in\mathbb{R}^{d\times d_u}\ \text{fixed},
\label{eq:fixed}
\end{equation}
whereas the MLP baseline replaces the fixed map with a trainable feed-forward transformation,
\begin{equation}
\mathbf{z}_v = \mathrm{MLP}_\theta(\mathbf{u}_v).
\label{eq:mlp}
\end{equation}
\begin{figure}
    \centering
    \includegraphics[width=1\linewidth]{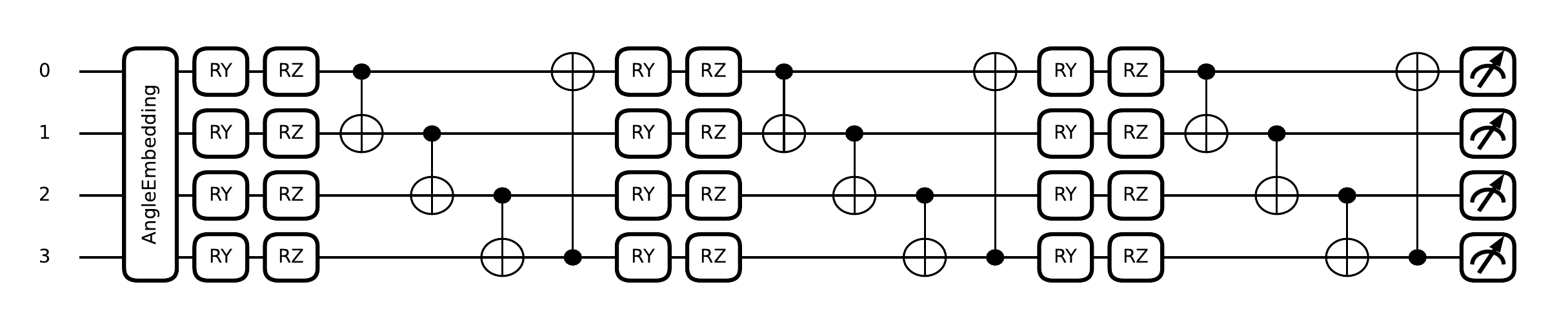}
    \caption{VQC used as the quantum embedding module. Angle embedding (Y rotations) is followed by $L_q$ entangling layers, each composed of per-qubit $R_Y$/$R_Z$ rotations and a ring of CNOT gates. Measurements include single-qubit $Z$ expectations.}
    \label{circuit}
\end{figure}
\paragraph{Circuit-defined embedding}
Angle-VQC generates per-node descriptors by applying a parameterized quantum circuit to an encoding of $\mathbf{u}_v$. First, $\mathbf{u}_v$ is mapped to $q$ rotation angles,
\begin{equation}
\boldsymbol{\varphi}_v = \mathbf{A}\mathbf{u}_v \in \mathbb{R}^{q},
\label{eq:angles}
\end{equation}
followed by an angle-encoding layer and $L_q$ variational layers (see Fig. \ref{circuit}), yielding the state
\begin{equation}
|\psi_v\rangle = U(\boldsymbol{\varphi}_v,\boldsymbol{\theta})\,|0\rangle^{\otimes q}.
\label{eq:vqc_state}
\end{equation}
A real-valued circuit descriptor is obtained by measuring a fixed observable set $\{\hat{O}_k\}_{k=1}^{m}$. In our experiments, we use single-qubit Pauli-$Z$ measurements on each qubit, i.e., $m=q$ and $\hat{O}_k = Z_k$:
\begin{equation}
s_{v,k}=\langle \psi_v|\hat{O}_k|\psi_v\rangle,\qquad k=1,\dots,m.
\label{eq:vqc_meas}
\end{equation}
This defines $\mathbf{s}_v=[s_{v,1},\dots,s_{v,m}]\in\mathbb{R}^{m}$, which is mapped to the shared node embedding dimension via Eq.~\eqref{eq:zv}. 
\subsection{Graph classifier and training protocol}
All embedding variants are evaluated with the same downstream classifier $f_\psi$ implemented as a GIN backbone \cite{xu2019how}. Starting from $\mathbf{H}^{(0)}=\mathbf{Z}$, message passing produces node states after $L$ layers:
\begin{equation}
\mathbf{H}^{(\ell)}=\mathrm{GIN}^{(\ell)}(\mathbf{H}^{(\ell-1)},G),\qquad \ell=1,\dots,L,
\label{eq:gin}
\end{equation}
followed by a permutation-invariant readout and linear head:
\begin{equation}
\mathbf{h}_G=\mathrm{Pool}(\mathbf{H}^{(L)}),\qquad 
\hat{\mathbf{y}}_G=\mathbf{W}_c\mathbf{h}_G+\mathbf{b}_c.
\label{eq:readout}
\end{equation}
For a dataset $\mathcal{D}=\{(G_i,y_i)\}_{i=1}^N$, parameters are optimized by minimizing cross-entropy on the training split,
\begin{equation}
\min_{\theta,\psi}\ \frac{1}{|\mathcal{D}_{\mathrm{train}}|}\sum_{(G,y)\in\mathcal{D}_{\mathrm{train}}}
\mathcal{L}\big(\hat{\mathbf{y}}_G,y\big),
\label{eq:loss}
\end{equation}
with early stopping based on validation Macro-F1 and reporting of test Accuracy, Macro-F1, and Macro Precision/Recall. Across all runs, the backbone $f_\psi$, stratified splits, optimization settings, early-stopping rule, and metric computation are kept fixed; consequently, performance differences reflect the embedding construction and its trainability.
\section{Results and Discussion}
\subsection{Experimental Setup}
We evaluate all embedding variants within a controlled graph-classification pipeline to isolate representation effects. All methods share the same downstream classifier (GIN backbone), identical stratified splits, identical optimization settings, and the same early-stopping criterion. 
The comparison includes two classical baselines (Fixed and MLP), avoiding more complex systems such as Transformers \cite{ying2021graphormer}, and a set of quantum-oriented embeddings. 
Among the quantum-oriented variants, Angle-VQC is the only circuit-defined embedding (node representations are generated by a parameterized quantum circuit using PennyLane's \texttt{default.qubit} simulator \cite{bergholm2018pennylane}), whereas QuOp, QWalkVec, and QPE are quantum-inspired constructions computed from graph operators and linear-algebraic transformations. 
Under this fixed protocol, performance differences primarily reflect the quality of the node representations produced by each embedding. 
The details for the evaluation setup are reported in Table \ref{tab:hyperparams}.

\begin{table}[htpb]
\centering
\caption{Hyperparameters and evaluation protocol used for all embedding variants (unless stated otherwise).}
\label{tab:hyperparams}
\scriptsize
\setlength{\tabcolsep}{3pt}
\renewcommand{\arraystretch}{1.08}
\begin{tabularx}{\columnwidth}{@{}>{\bfseries}l X@{}}
\toprule
Category & Setting \\
\midrule
Datasets &
TU: IMDB-BINARY, IMDB-MULTI, MUTAG, PROTEINS, ENZYMES; QM9 (classification via binning). \\
Splits &
Stratified 80/10/10 (train/val/test), seed $=7$. \\
Positional encoding &
Laplacian eigenvector PE, $k=8$, stored as \texttt{pe}. \\
Embedding output &
Node embedding dim $d=32$ (all methods). \\
Backbone &
GIN (3 layers), hidden dim $64$, global mean pooling; MLP head $64\!\rightarrow\!64\!\rightarrow\!C$; dropout $0.2$. \\
Optimization &
Adam; lr $10^{-3}$; weight decay $0$; max epochs $30$. \\
Early stopping &
Monitor val Macro-F1; patience $7$; select best checkpoint. \\
Batch size&
$16$. \\
QM9 labeling &
Target index $0$; 2 bins; quantile binning; max graphs $5000$. \\
QWalkVec &
$t=32$, $w_p=0.5$, $w_q=4.0$; stored as \texttt{qwalkvec}. \\
QPE &
Times $\{0.5, 1.0, 2.0\}$; anchors $8$ \\
\bottomrule
\end{tabularx}
\end{table}

Table~\ref{tab:embed_perf} reports test-set Accuracy, Macro-F1, and Macro-P/R for all embedding variants under this protocol, and the following analysis discusses the observed dataset-dependent trends and the conditions under which quantum-oriented embeddings provide improvements over strong classical baselines.
\begin{table}[t]
\centering
\caption{Performance comparison across embedding variants (test set). Best values are highlighted per dataset (Acc and Macro-F1).}
\label{tab:embed_perf}
\small
\begin{adjustbox}{max width=\columnwidth}
\begin{tabular}{@{}c l c c c c@{}}
\toprule
\textbf{Data} & \textbf{Method} & \textbf{Tr.} & \textbf{Acc} & \textbf{F1} & \textbf{P/R} \\
\midrule

\multirow{8}{*}{\rotatebox{90}{\textbf{IMDB}}}
& Fixed     & N & 0.71   & 0.7097 & 0.7237/0.7202 \\
& MLP       & Y & \best{0.72}   & \best{0.7172} & 0.7172/0.7172 \\
& Angle-VQC & Y & 0.60   & 0.4667 & 0.7895/0.5556 \\
& QuOp    & N & 0.62   & 0.6200 & 0.6263/0.6263 \\
& QuOp*      & Y & 0.65   & 0.6491 & 0.6500/0.6515 \\
& QWalkVec  & N & 0.67   & 0.6576 & 0.7520/0.6959 \\
& QWalkVec* & Y & 0.68   & 0.6716 & 0.6776/0.6707 \\
& QPE       & N & 0.63   & 0.6297 & 0.6322/0.6333 \\
\midrule

\multirow{8}{*}{\rotatebox{90}{\textbf{IMDB-MULTI}}}
& Fixed     & N & 0.4267 & 0.3914 & 0.4033/0.4267 \\
& MLP       & Y & 0.4400 & 0.4132 & 0.4232/0.4400 \\
& Angle-VQC & Y & 0.4467 & 0.4024 & 0.4115/0.4467 \\
& QuOp    & N & \best{0.5067} & \best{0.4662} & 0.5280/0.5067 \\
& QuOp*      & Y & 0.4733 & 0.4660 & 0.4703/0.4733 \\
& QWalkVec  & N & 0.4267 & 0.4299 & 0.4361/0.4267 \\
& QWalkVec* & Y & 0.4533 & 0.4244 & 0.4370/0.4533 \\
& QPE       & N & 0.4200 & 0.4058 & 0.4028/0.4200 \\
\midrule

\multirow{8}{*}{\rotatebox{90}{\textbf{MUTAG}}}
& Fixed     & N & 0.7895 & 0.7841 & 0.8000/0.8462 \\
& MLP       & Y & 0.8421 & 0.8348 & 0.8333/0.8846 \\
& Angle-VQC & Y & 0.8947 & 0.8782 & 0.8782/0.8782 \\
& QuOp    & N & 0.8421 & 0.8348 & 0.8333/0.8846 \\
& QuOp*      & Y & 0.8421 & 0.8348 & 0.8333/0.8846 \\
& QWalkVec  & N & 0.4211 & 0.3942 & 0.6765/0.5769 \\
& QWalkVec* & Y & \best{0.9474} & \best{0.9360} & 0.9643/0.9167 \\
& QPE       & N & 0.8421 & 0.7816 & 0.9062/0.7500 \\
\midrule

\multirow{8}{*}{\rotatebox{90}{\textbf{PROTEINS}}}
& Fixed     & N & 0.7679 & 0.7617 & 0.7596/0.7658 \\
& MLP       & Y & 0.7589 & 0.7360 & 0.7606/0.7292 \\
& Angle-VQC & Y & 0.6786 & 0.5714 & 0.7813/0.6036 \\
& QuOp    & N & 0.7054 & 0.6647 & 0.7083/0.6625 \\
& QuOp*      & Y & 0.6518 & 0.6365 & 0.6370/0.6360 \\
& QWalkVec  & N & 0.7321 & 0.7114 & 0.7247/0.7068 \\
& QWalkVec* & Y & \best{0.7768} & \best{0.7630} & 0.7708/0.7587 \\
& QPE       & N & 0.7321 & 0.6973 & 0.7408/0.6922 \\
\midrule

\multirow{8}{*}{\rotatebox{90}{\textbf{ENZYMES}}}
& Fixed     & N & 0.2667 & 0.2160 & 0.1981/0.2667 \\
& MLP       & Y & 0.2500 & 0.2242 & 0.2177/0.2500 \\
& Angle-VQC & Y & 0.1500 & 0.1317 & 0.1402/0.1500 \\
& QuOp    & N & 0.1833 & 0.1871 & 0.2026/0.1833 \\
& QuOp*      & Y & 0.1833 & 0.1658 & 0.1703/0.1833 \\
& QWalkVec  & N & \best{0.3000} & \best{0.2827} & 0.2805/0.3000 \\
& QWalkVec* & Y & 0.1833 & 0.1556 & 0.1627/0.1833 \\
& QPE       & N & 0.2500 & 0.2066 & 0.1793/0.2500 \\
\midrule

\multirow{8}{*}{\rotatebox{90}{\textbf{QM9}}}
& Fixed     & N & 0.8300 & 0.8300 & 0.8300/0.8300 \\
& MLP       & Y & 0.8120 & 0.8107 & 0.8211/0.8120 \\
& Angle-VQC & Y & 0.8060 & 0.8053 & 0.8102/0.8060 \\
& QuOp    & N & 0.7240 & 0.7152 & 0.7557/0.7240 \\
& QuOp*      & Y & 0.7520 & 0.7502 & 0.7593/0.7520 \\
& QWalkVec  & N & 0.7200 & 0.7162 & 0.7325/0.7200 \\
& QWalkVec* & Y & \best{0.8520} & \best{0.8518} & 0.8543/0.8520 \\
& QPE       & N & 0.7780 & 0.7746 & 0.7956/0.7780 \\
\bottomrule
\end{tabular}
\end{adjustbox}
\end{table}
\subsection{Dataset-wise performance patterns}
Performance varies systematically across dataset families. On \textbf{IMDB}, the strongest results are obtained by the classical baselines (MLP: Acc $=0.72$, Macro-F1 $=0.7172$; Fixed: Acc $=0.71$, Macro-F1 $=0.7097$). In this attribute-scarce regime, where node features are constructed to ensure $\mathbf{x}\neq\varnothing$ but remain intentionally minimal, additional embedding complexity does not consistently translate into improved class-balanced performance. Angle-VQC underperforms sharply (Acc $=0.60$, Macro-F1 $=0.4667$) and exhibits a pronounced Macro-P/R mismatch, indicating uneven class-wise behavior under the current training budget. QuOp/QWalkVec/QPE remain competitive but do not surpass the strongest baselines on IMDB.

On \textbf{IMDB-MULTI}, the ranking changes and quantum-inspired structures become beneficial. QuOp achieves the best overall result (Acc $=0.5067$, Macro-F1 $=0.4662$), improving over Fixed/MLP (Acc $\approx 0.43$--$0.44$, Macro-F1 $\approx 0.39$--$0.41$). The trainable QuOp* does not exceed QuOp (similar Macro-F1 but lower accuracy), suggesting that the operator-induced inductive bias is effective on its own, while additional trainable capacity is not consistently converted into better generalization in this setting. Angle-VQC attains moderate accuracy but does not improve Macro-F1 relative to QuOp, consistent with non-uniform class performance in the multi-class regime.

For structure-driven benchmarks, quantum-oriented embeddings yield the clearest gains over baselines. 

On \textbf{MUTAG}, QWalkVec* dominates with Acc $=0.9474$ and Macro-F1 $=0.9360$, clearly exceeding all alternatives; Angle-VQC is also strong (Acc $=0.8947$, Macro-F1 $=0.8782$), while Fixed/MLP/QuOp* cluster around Acc $\approx 0.84$. The contrast within QWalkVec is informative: the non-trainable variant collapses (Acc $=0.4211$) while QWalkVec* excels, indicating that walk-derived descriptors require a task-aligned projection to become effective. 

On \textbf{PROTEINS}, QWalkVec* yields highest Macro-F1 (Acc $=0.7768$, Macro-F1 $=0.7630$), providing a modest improvement over Fixed (Acc $=0.7679$, Macro-F1 $=0.7617$). 

On \textbf{ENZYMES}, absolute scores remain low across all methods, but QWalkVec achieves the strongest result among evaluated variants (Acc $=0.3000$, Macro-F1 $=0.2827$), improving over Fixed (Acc $=0.2667$, Macro-F1 $=0.2160$), indicating that richer multi-hop structural descriptors are still useful in a harder multi-class setting. 

Finally, on \textbf{QM9} (binary classification via target binning), QWalkVec* again provides the best performance (Acc $=0.8520$, Macro-F1 $=0.8518$), improving over Fixed (Acc $=0.8300$), while QuOp* and QPE are weaker in this configuration.

\subsection{Where the quantum-oriented benefit comes from}
Two factors explain when quantum-oriented embeddings improve over baselines. First, the advantage is strongest when labels correlate with \emph{multi-hop structure} rather than weak constructed node attributes, as reflected by the consistent gains of QWalkVec* on MUTAG and QM9 and its improvements on PROTEINS and ENZYMES. Second, the effect depends on how the quantum-oriented signal is integrated into the representation space. QWalkVec shows that a learned projection can be essential: without it, walk-derived descriptors may be misaligned with the downstream loss (MUTAG collapse), whereas with a trainable mapping, the same structural signal becomes highly predictive. In contrast, QuOp illustrates that a fixed operator-based inductive bias can already be effective on IMDB-MULTI, while adding trainable capacity (QuOp*) does not reliably improve outcomes under a fixed optimization budget.
\subsection{Hard regimes and failure modes suggested by ENZYMES.}
ENZYMES stands out as uniformly difficult under the present configuration: all methods obtain low Macro-F1 and Accuracy (best Acc. $=0.30$).
Given that the backbone and training protocol are held fixed, this suggests that the tested embeddings, under the current backbone capacity and training budget, are not sufficient to elicit strong performance on this multi-class setting.
Methodologically, this motivates targeted ablations in regimes where gains are modest, for example, varying the embedding fusion strategy, revisiting the scaling between base inputs and dynamics-derived features, or increasing shared-backbone capacity while maintaining the same controlled comparison framework.

\subsection{Computational characteristics implied by the implementations.}
The embedding families differ substantially in their computational profile as implemented here, which matters when interpreting practical trade-offs. QuOp requires per-node construction of an ego-neighborhood, padding to a power-of-two dimension, and a matrix exponential to obtain a unitary before measurement-style summaries; the cost therefore scales with both neighborhood size and the chosen qubit budget. 
QPE requires a per-graph eigendecomposition to evaluate $U(t)=\exp(-iHt)$, after which multiple time points can be evaluated cheaply but still depend on dense linear algebra on the graph size.
QWalkVec builds a directed-edge state space and iterates a coined-walk update for a fixed number of steps, accumulating node probabilities over time; its cost is dominated by the walk horizon and the number of directed edges. 
These differences motivated a conservative batch size and early stopping strategy in the shared training protocol, ensuring that comparisons remain feasible across datasets while keeping the downstream model fixed.

\subsection{Trainability and stability}
Trainability is not uniformly beneficial and should be treated as a controlled design axis. QWalkVec exhibits the strongest dependence on trainability (essential on MUTAG and beneficial on PROTEINS/QM9), whereas QuOp* shows that frozen representations can be preferable (IMDB-MULTI). Angle-VQC further illustrates dataset sensitivity \cite{mcclean2018barren}: it performs strongly on MUTAG but fails to generalize on IMDB and degrades on PROTEINS, indicating that circuit-defined embeddings can be more sensitive to the data regime and to optimization constraints.

\subsection{Discussion and takeaways}
Our results support three takeaways for embedding selection in graph classification. 
\begin{itemize}
    \item \textbf{Quantum-oriented embeddings can yield clear gains} in structure-driven regimes: walk-based quantum-inspired embeddings (QWalkVec*) provide the best performance on MUTAG and QM9 and remain competitive or best on PROTEINS and ENZYMES, demonstrating that quantum-motivated structural descriptors can outperform strong baselines under a fixed backbone. 
    \item \textbf{The advantage is dataset-dependent}: on attribute-scarce social graphs (IMDB), classical baselines remain strongest, suggesting that the limiting factor is the weakness of node-level signal rather than embedding expressiveness. 
    \item \textbf{Optimization stability matters as much as inductive bias}: trainability can be essential (QWalkVec) or neutral/negative (QuOp on IMDB-MULTI), so fair comparisons should report both frozen and trainable variants and rely on Macro-F1/Macro-P/R to expose class-wise failure modes that accuracy alone can mask.
\end{itemize}

\section{Conclusion}

This study isolates the effect of the node-embedding stage in graph classification by evaluating classical and quantum-oriented node representations under a fixed backbone (GIN), fixed splits, and a shared optimization protocol. 
The results show a consistent pattern: embeddings that explicitly inject multi-hop structural information can improve graph-level performance when labels are structure-driven, while attribute-scarce social graphs remain well served by simple baselines built from lightweight structural and positional cues.

Among the quantum-oriented variants, walk-derived descriptors are the most reliable source of gains, but only when their outputs are mapped into a task-aligned representation space: the strongest improvements appear when the walk signal is coupled to a learnable projection, whereas the corresponding frozen configuration can underperform sharply. 
Operator-based encodings exhibit a different profile, providing competitive performance without always benefiting from additional trainable capacity, which suggests that the inductive bias of the construction itself can be the dominant factor under a fixed training budget. 
Circuit-defined embeddings show clear dataset sensitivity, indicating that trainability and optimization stability remain practical constraints when the embedding is produced by a variational circuit.

Overall, the benchmark supports a simple takeaway for practitioners: when the target depends on graph structure beyond immediate neighborhoods, quantum-inspired dynamics can be a useful way to enrich node representations, but their benefit depends on how the resulting descriptors are interfaced with the downstream learner. 
A natural next step is to stress-test these findings under larger training budgets and multi-seed evaluation, and to study whether the same embedding choices remain favorable when paired with backbones that are explicitly designed for long-range dependencies.

\section*{Acknowledgment}
This work was supported in part by the NYUAD Center for Quantum and Topological Systems (CQTS), funded by Tamkeen under the NYUAD Research Institute grant CG008.

\bibliographystyle{IEEEtran}

\bibliography{refs}

\end{document}